\documentclass{article}

\usepackage{PRIMEarxiv}

\usepackage[utf8]{inputenc} 
\usepackage[T1]{fontenc}    
\usepackage{hyperref}       
\usepackage{url}            
\usepackage{booktabs}       
\usepackage{amsfonts}       
\usepackage{nicefrac}       
\usepackage{microtype}      
\usepackage{lipsum}
\usepackage{fancyhdr}       
\usepackage{float}
\usepackage{graphicx}       
\usepackage{placeins}
\graphicspath{{media/}}     

\title{Technical Report on classification of literature related to children speech disorder
}

\author{
  Ziang Wang \\
  Australian National University \\
  \texttt{ziang.wang@anu.edu.au} \\
   \And
  Amir Aryani \\
  Swinburne University of Technology \\
  \texttt{aaryani@swin.edu.au} \\
}

\begin{document}
\thispagestyle{empty}
\maketitle
\pagestyle{plain}

\begin{abstract}
This technical report presents a natural language processing (NLP)-based approach for systematically classifying scientific literature on childhood speech disorders. We retrieved and filtered 4,804 relevant articles published after 2015 from the PubMed database using domain-specific keywords. After cleaning and pre-processing the abstracts, we applied two topic modeling techniques—Latent Dirichlet Allocation (LDA) and BERTopic—to identify latent thematic structures in the corpus. Our models uncovered 14 clinically meaningful clusters, such as infantile hyperactivity and abnormal epileptic behavior. To improve relevance and precision, we incorporated a custom stop word list tailored to speech pathology. Evaluation results showed that the LDA model achieved a coherence score of 0.42 and a perplexity of -7.5, indicating strong topic coherence and predictive performance. The BERTopic model exhibited a low proportion of outlier topics (<20\%), demonstrating its capacity to classify heterogeneous literature effectively. These results provide a foundation for automating literature reviews in speech-language pathology.
\end{abstract}


\section{Introduction}

Childhood speech disorders are a prevalent and serious condition that affect language comprehension, expression, fluency, and other aspects of communication. These disorders not only impair children's academic performance and social interaction, but also contribute to long-term psychological and behavioral issues, placing considerable burdens on families and society.

According to the National Institute on Deafness and Other Communication Disorders, approximately 8\% to 9\% of young children experience speech sound difficulties~\cite{nidcd_stats}, which may contribute to long-term challenges in reading and language development. In particular, a recent population-based study found that 3.6\% of 8-year-old children exhibit persistent speech sound disorders~\cite{wren2016prevalence}. These findings underscore the high prevalence and potential long-term impact of speech impairments during early development.

Beyond individual challenges, childhood speech disorders incur significant socioeconomic costs. Impaired language abilities limit children's educational outcomes, social integration, and emotional development. Studies have linked early language delays with increased risks of academic failure, social withdrawal, and mental health conditions, which in turn reduce employment prospects and raise long-term healthcare demands~\cite{cambridge_econ_impact}.

Despite advancements in speech pathology research, several gaps remain unresolved. The field is inherently interdisciplinary—encompassing audiology, pediatrics, neurology, psychology, and rehabilitation science—yet suffers from limited cross-disciplinary integration. This fragmentation hampers comprehensive understanding and slows the translation of research findings into clinical practice.

To address these challenges, we propose a natural language processing (NLP)-based framework to systematically classify literature on children's speech disorders. By applying topic modeling techniques to a curated corpus from PubMed, we aim to identify latent thematic structures and support knowledge synthesis across disciplines.

The main contributions of this work are as follows:
\begin{itemize}
    \item We present a reproducible pipeline for automated literature classification in speech pathology.
    \item We apply and compare two topic modeling approaches—Latent Dirichlet Allocation (LDA) and BERTopic—on real-world data.
    \item We provide a quantitative evaluation of topic coherence and model effectiveness.
\end{itemize}

\begin{figure}[htbp]
    \centering
    \includegraphics[width=0.8\textwidth]{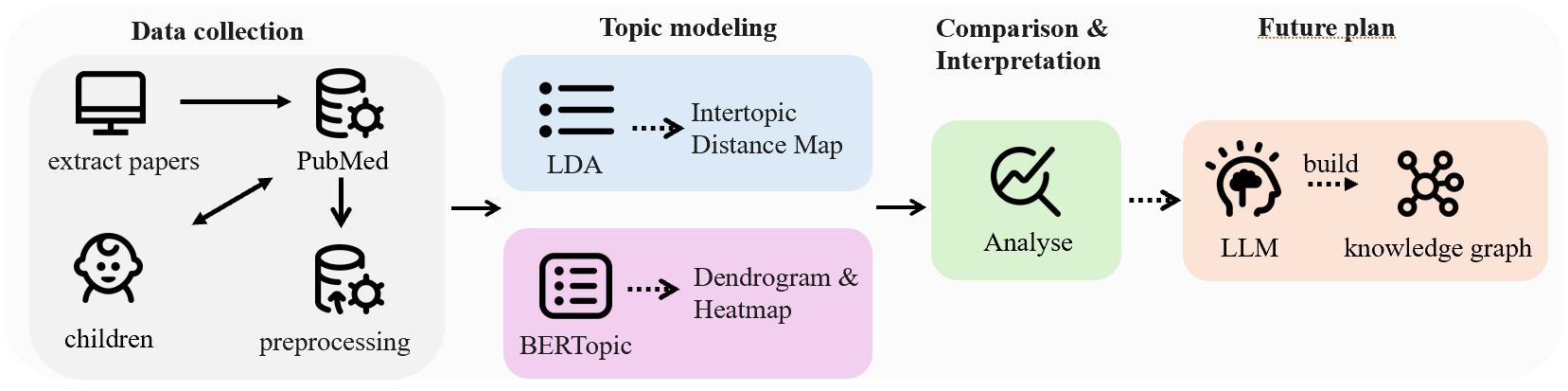}
    \caption{Overview of the topic modeling workflow using LDA and BERTopic}
    \label{fig:workflow}
\end{figure}

To provide a clearer overview of the analytical process undertaken in this study, Figure~\ref{fig:workflow} illustrates the complete workflow from data preparation to topic evaluation.

\FloatBarrier
\section{Data Collection}

To construct a relevant corpus, literature related to childhood speech disorders published between 2015 and 2025 was retrieved from the PubMed database. The search used the following keywords: \textit{stuttering}, \textit{stammering}, \textit{speech disorder}, \textit{communication disorder}, \textit{language disorder}, \textit{tempering}, and \textit{temperament}. Boolean operators (e.g., AND, OR) were used to broaden the search scope. The search was conducted via the Entrez Programming Utilities (E-utilities) through Biopython’s \texttt{Entrez} module~\cite{cock2009biopython}.

The query results were parsed and exported into a structured CSV file including the following metadata: PubMed ID (PMID), article title, authors, year of publication, journal name, abstract, and DOI link. A total of 4,804 records were retrieved.

The collected dataset underwent multiple preprocessing steps. First, articles not written in English were removed, resulting in the exclusion of 250 records. Then, to ensure relevance to childhood populations, abstracts and titles were filtered using a predefined set of child-related keywords, including \textit{child}, \textit{infant}, \textit{toddler}, \textit{pediatric}, \textit{adolescent}, and \textit{preschool}. Articles lacking these terms in either the title or abstract were excluded, leading to the removal of 539 records. Keyword filtering was case-insensitive and performed using regular expression matching to account for plural forms and spelling variants.

After filtering, a final dataset of 4,015 English-language publications specifically focusing on children’s speech disorders was obtained for further analysis. Table~\ref{tab:data-filter} shows the number of documents that were removed and left during data pre-processing.

\begin{table}[h]
\centering
\caption{Data Filtering Statistics}
\begin{tabular}{lrr}
\toprule
\textbf{Step} & \textbf{Removed} & \textbf{Remaining} \\
\midrule
Initial retrieval & -- & 4,804 \\
Non-English articles removed & 250 & 4,554 \\
Non-child-related articles removed & 539 & 4,015 \\
\bottomrule
\end{tabular}
\label{tab:data-filter}
\end{table}

\FloatBarrier
\section{Clustering Methods}

To uncover latent themes within the collected literature, this study employed unsupervised topic modeling techniques. Topic modeling enables the identification of semantically coherent clusters of words—referred to as “topics”—which in turn help to characterize document contents at a thematic level. In this research, two distinct approaches were used: the traditional Latent Dirichlet Allocation (LDA) and the more recent BERTopic model. This dual-method strategy allowed for the comparison between probabilistic topic modeling based on word frequency and a transformer-based embedding approach capable of capturing contextual semantics.

\subsection{Latent Dirichlet Allocation (LDA)}

LDA~\cite{blei2003latent} is a generative model that assumes each document in a corpus is composed of multiple latent topics, and that each topic corresponds to a distribution over words. The model estimates the posterior topic distributions by analyzing word co-occurrence patterns, and assigns probabilistic topic weights to each document. 

In preparation for modeling, a series of preprocessing steps were undertaken. The abstracts of 4,015 papers were first cleaned by eliminating missing entries and non-informative records. The Natural Language Toolkit (NLTK) was used to remove standard English stopwords, which were further supplemented by a domain-specific list including high-frequency but non-informative terms such as “result”, “method”, and “patient”. A bag-of-words representation was constructed using the Gensim library. Tokens that appeared in fewer than five documents or in more than 95\% of the corpus were discarded to reduce noise.

LDA models were trained with varying numbers of topics, ranging from 10 to 30. For each configuration, two key evaluation metrics were computed: perplexity and topic coherence. Perplexity quantifies how well the model predicts a sample of held-out data, with lower values indicating better generalization. In contrast, topic coherence measures the semantic similarity among the top terms within each topic, with higher values suggesting better interpretability. After analyzing both metrics, the model with 14 topics was selected as optimal. This configuration struck a balance between minimizing perplexity and maximizing coherence.

\subsection{BERTopic}

To complement LDA’s probabilistic modeling, this study also applied BERTopic~\cite{grootendorst2022bertopic}, a clustering-based technique that leverages document embeddings derived from pretrained language models. Unlike LDA, which relies on raw word counts, BERTopic captures deeper semantic representations by mapping documents into a high-dimensional embedding space.

For this analysis, document embeddings were generated using the \texttt{all-MiniLM-L6-v2} model from the SentenceTransformers library. Text preprocessing steps mirrored those used for LDA, with additional adjustments to accommodate the requirements of vector-based representations. A customized vectorizer was applied to remove stopwords, extract frequent bigrams, and eliminate rare or overly common tokens. Following this, dimensionality reduction was performed using UMAP, and clusters were identified via HDBSCAN, an algorithm well-suited to handling variable-density clusters and noise in high-dimensional spaces.

To evaluate the quality of topics, several analyses were conducted. These included examining the distribution of documents across topics, identifying the proportion of outlier documents (those that could not be confidently assigned to any cluster), and inspecting a heatmap of cosine similarities between topic embeddings. Based on these diagnostics, post-processing adjustments were made to merge semantically similar clusters and reassign ambiguous documents, thereby improving the clarity and separation of topics.

\section{Discussion}

\subsection{Topic Modeling with LDA}

\begin{figure}[htbp]
    \centering
    \includegraphics[width=0.75\linewidth]{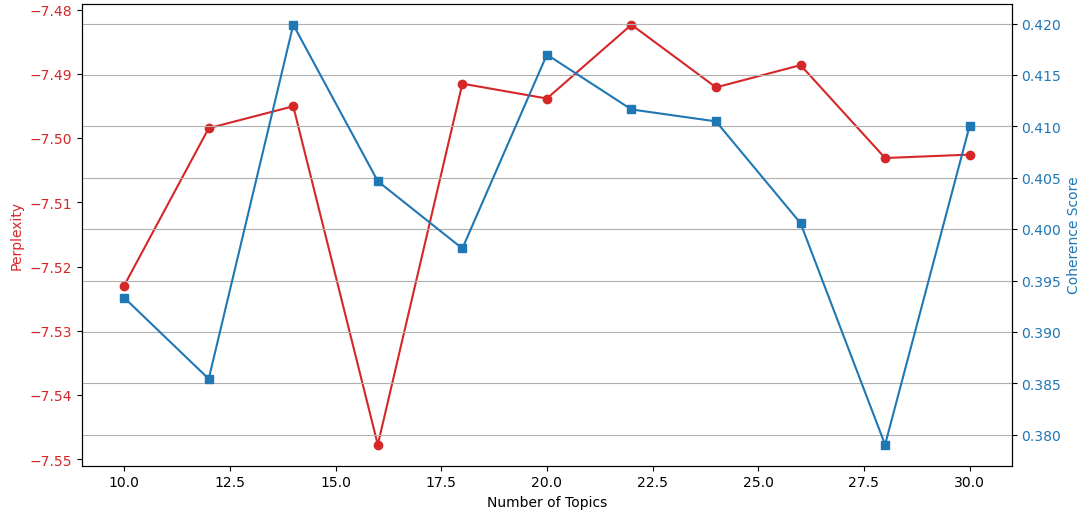}
    \caption{Optimal Number of Topics Evaluation for LDA}
    \label{fig:lda-eval}
\end{figure}

Latent Dirichlet Allocation (LDA) was initially applied to uncover latent semantic structures within the corpus. The number of topics was varied between 10 and 30 to identify an optimal model configuration. Two evaluation metrics guided model selection: topic coherence (using the $C_v$ measure) and perplexity. While perplexity evaluates generalizability by assessing how well the model predicts unseen texts, coherence—reflecting the semantic relatedness of top topic words—was prioritized due to its stronger alignment with human interpretability. As shown in Figure~\ref{fig:lda-eval}, the coherence score peaked when 14 topics were used. Beyond this point, performance declined slightly, implying that the inclusion of additional topics introduced semantic redundancy rather than finer granularity. 

\begin{figure}[htbp]
    \centering
    \includegraphics[width=0.5\linewidth]{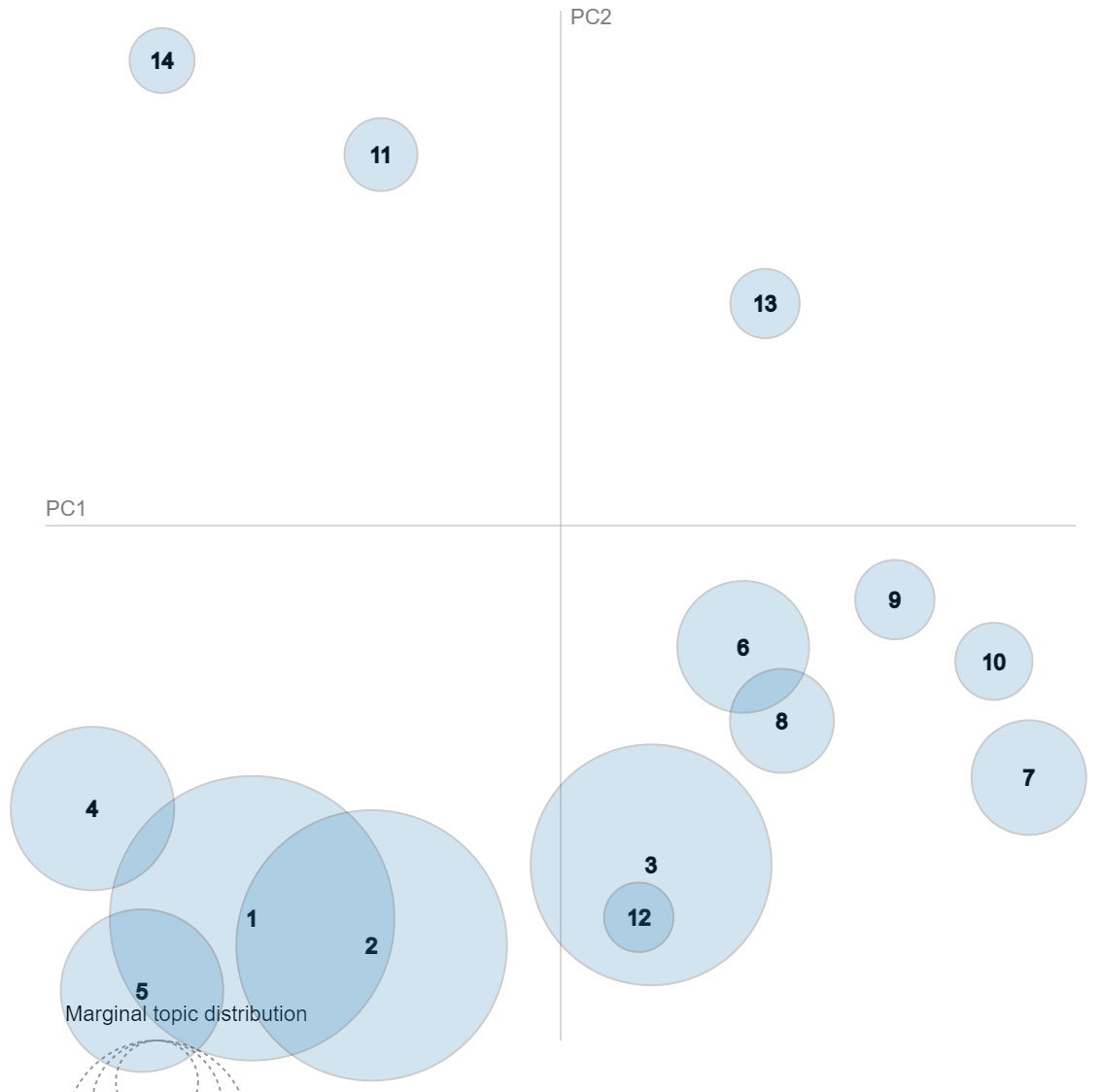}
    \caption{Intertopic Distance Map for LDA}
    \label{fig:lda-intertopic}
\end{figure}

The extracted topics spanned a range of relevant domains, including developmental language disorder (DLD), attention-deficit/hyperactivity disorder (ADHD), emotion regulation, and parental stress. Nevertheless, the intertopic distance map (Figure~\ref{fig:lda-intertopic}) suggested that certain topics overlapped in semantic space, potentially indicating insufficient topic distinctiveness. This could be attributed to the limitations of LDA’s reliance on co-occurrence patterns and the constraints imposed by the preprocessing pipeline.

\FloatBarrier
\subsection{Topic Modeling with BERTopic}

BERTopic was subsequently employed as a complementary approach that leverages contextualized embeddings and density-based clustering. Unlike LDA, BERTopic integrates transformer-based sentence representations (BERT), dimensionality reduction via UMAP, and clustering through HDBSCAN, followed by a CountVectorizer-based representation. This architecture is more sensitive to hyperparameter choices, particularly those governing minimum cluster size and distance metrics. Parameter tuning focused on optimizing UMAP and CountVectorizer settings, with the final configuration being: \texttt{ngram\_range=(1,3)}, \texttt{min\_df=5}, \texttt{max\_df=0.95}, \texttt{n\_neighbors=8}, \texttt{n\_components=8}, \texttt{min\_dist=0.1}, and \texttt{metric='manhattan'}. Under these settings, the outlier rate was maintained below 20\%, indicating a reasonably stable clustering outcome. 

\begin{figure}[htbp]
    \centering
    \includegraphics[width=0.75\linewidth]{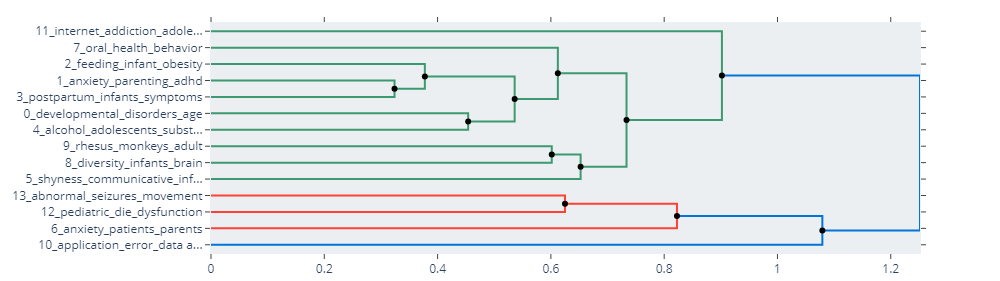}
    \caption{Hierarchical Clustering for BERTopic}
    \label{fig:bertopic-tree}
\end{figure}

\begin{figure}[htbp]
    \centering
    \includegraphics[width=0.75\linewidth]{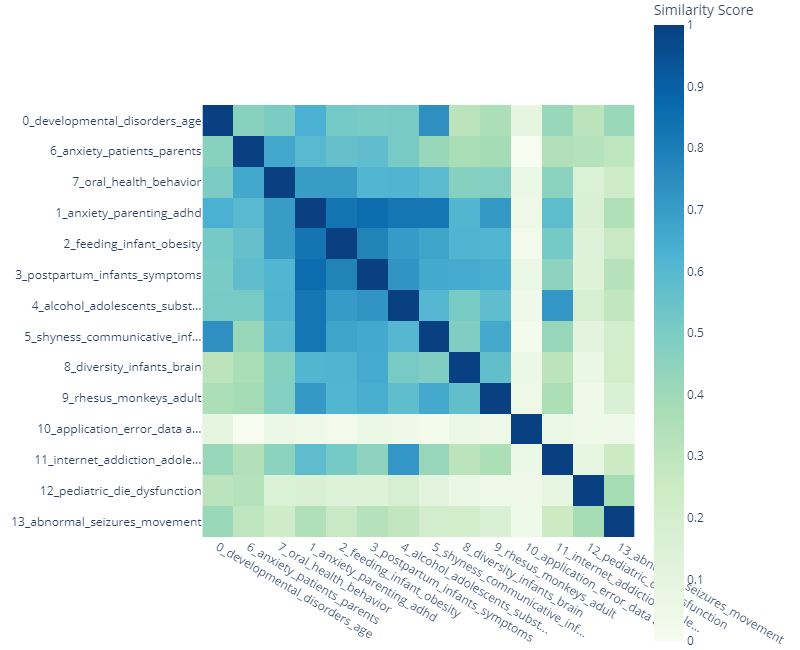}
    \caption{Heat Map for BERTopic}
    \label{fig:bertopic-heatmap}
\end{figure}

The resulting 14 topics included themes such as internet addiction, maternal-infant bonding, and pediatric emergency care. A dendrogram (Figure~\ref{fig:bertopic-tree}) and a heatmap (Figure~\ref{fig:bertopic-heatmap}) were generated to visualize topic separation and inter-topic relationships. Although BERTopic offered improved flexibility and supported hierarchical topic structures, some clusters exhibited notable similarity, suggesting opportunities for further refinement through more targeted parameter tuning.

\FloatBarrier
\subsection{Comparison and Interpretation}

A structured comparison of LDA and BERTopic is presented in Table~\ref{tab:model-comparison}. 

\begin{table}[htbp]
\centering
\caption{Comparison between LDA and BERTopic}
\label{tab:model-comparison}
\begin{tabular}{@{} l p{5cm} p{5cm} @{}}
\toprule
\textbf{Aspect} & \textbf{LDA} & \textbf{BERTopic} \\
\midrule
Topic Representation & Bag-of-words, keyword driven & Semantic embedding with contextual meaning \\
Hierarchy Support & Not supported & Supported via dendrogram \\
Interpretability & Clear keyword themes & Some ambiguous labels \\
Topic Redundancy & Moderate (as shown in intertopic map) & Less, but sensitive to parameters \\
Evaluation Metrics & Coherence: 0.42, Perplexity: -7.495 & Outlier ratio: < 20\%, Heatmap similarity noted \\
Computational Cost & Low & High \\
\bottomrule
\end{tabular}
\end{table}

The topics produced by LDA—such as those centered around developmental language disorder and ADHD—tended to exhibit high interpretability, primarily due to the model’s emphasis on frequently co-occurring keywords. However, overlap between several topic domains was noted, reflecting the inherent limitations of the bag-of-words approach in capturing nuanced semantic differences. BERTopic, by contrast, employed contextual embeddings to detect finer-grained distinctions across texts. While this enabled a more differentiated thematic structure, it occasionally resulted in topic labels that were less immediately transparent. One of BERTopic’s strengths lay in its support for hierarchical exploration through dendrogram visualizations—an analytical layer not available in LDA—that allows researchers to examine topic relationships across multiple levels of abstraction. Nevertheless, this enhanced representational capacity came at the cost of increased computational demands and heightened sensitivity to hyperparameter settings. 

These observations underscore that model choice should be informed by analytical priorities: LDA remains a viable option for fast, interpretable topic generation under resource constraints, whereas BERTopic is better suited for in-depth semantic exploration when interpretability can be balanced with complexity.

\subsection{Limitations and Future Work}

Despite their respective strengths, both LDA and BERTopic exhibited challenges related to topic redundancy and semantic overlap. In the case of LDA, closely related topics occasionally clustered together, reflecting the limitations of co-occurrence-based representations. BERTopic, though context-aware, sometimes produced clusters with high internal similarity, suggesting suboptimal separation. These issues may stem in part from preprocessing limitations, such as the absence of domain-specific named entity recognition (NER), and from the structural assumptions of each model. Furthermore, both approaches encountered difficulties in assigning clinically precise topic labels—an essential requirement for sensitive domains such as speech-language pathology. 

Future work could address these limitations by incorporating biomedical or psychological ontologies (e.g., UMLS) into the preprocessing pipeline, implementing supervised or semi-supervised topic labeling methods, and exploring dynamic topic modeling frameworks to capture temporal shifts in discourse. Such enhancements may improve both the interpretability and the clinical relevance of topic modeling in specialized corpora.

\section{Data and Code}
In this section, we describe the open-access dataset and source code used in the topic modeling analysis. All data and source code used in this study are accessible via Zenodo~\cite{zenodo_record}.

\subsection{Data}
Two datasets are provided in CSV format. The first, \texttt{speech\_disorders\_children\_2016\_present.csv}, contains metadata of research papers related to childhood speech and language disorders, extracted from PubMed for the period 2016 to present. Articles not written in English were excluded. A data cleaning process was conducted to ensure relevance to childhood topics, and the final curated dataset is saved as \texttt{speech\_disorders\_cleaned.csv}.

\subsection{Code}
All Python scripts and Jupyter notebooks used in the study are included:
\begin{itemize}
    \item \texttt{pubmed\_fetcher.py}: A Python script using Entrez API to retrieve relevant PubMed articles.
    \item \texttt{data\_processing.ipynb}: Notebook for preprocessing, including filtering non-English papers and removing irrelevant entries.
    \item \texttt{topic\_modeling\_LDA.ipynb}: Implements LDA-based topic modeling on the cleaned dataset.
    \item \texttt{topics\_number\_LDA.ipynb}: Evaluates LDA models using coherence and perplexity to determine the optimal topic number.
    \item \texttt{topic\_modeling\_BERTopic.ipynb}: Applies BERTopic to identify latent themes using contextual embeddings.
    \item \texttt{topics\_number\_BERTopic\_14.ipynb}: Assesses topic distinctiveness using dendrogram and heatmap visualizations for BERTopic results.
\end{itemize}
All dependencies are listed in the \texttt{requirements.txt} file, and the project structure is documented in the \texttt{README.md}. Instructions for reproducing the analysis are included in the repository.

\section{Conclusion}

This study proposed a computational framework for the unsupervised classification of biomedical literature on childhood speech disorders. By applying and comparing two topic modeling techniques—Latent Dirichlet Allocation (LDA) and BERTopic—on a curated dataset from PubMed, the study demonstrated several methodological contributions. Specifically, it established a reproducible pipeline for large-scale literature analysis in speech pathology, employed multiple evaluation metrics (topic coherence, perplexity, and proportion of incoherent topics), and generated interpretable topic clusters relevant to clinical and developmental speech research.

The analysis revealed several dominant themes, such as early diagnosis, intervention strategies, neural correlates, and comorbid developmental conditions, highlighting the diverse research focus within the domain. Furthermore, the comparison of LDA and BERTopic provided insights into the trade-offs between interpretability and semantic richness, with BERTopic offering superior cluster coherence in most cases.

Future work can expand this framework by integrating multilingual and multidisciplinary databases (e.g., Embase, LILACS) to capture regional and linguistic variations in research. Additionally, incorporating dynamic topic tracking mechanisms may facilitate the monitoring of emerging therapies and evolving clinical guidelines. Finally, coupling topic analysis with patient-reported outcomes and clinical trial metadata may enable a more patient-centered understanding of speech disorder research.

Overall, this study demonstrates the feasibility and value of combining natural language processing with biomedical literature mining, providing a foundation for further computational exploration in evidence synthesis and clinical informatics.

\section*{Acknowledgments}
This work was supported in part by the Research Graph Foundation as part of the internship program on open research infrastructure and knowledge mining. We would like to thank the Foundation for providing access to technical resources and expert guidance throughout the project. We are also grateful to Dr. Amir Aryani from Swinburne University of Technology for his supervision and valuable insights during the development of this report.

\bibliographystyle{unsrt}  
\bibliography{references}

\begin{thebibliography}{1}

\bibitem{nidcd_stats}
{National Institute on Deafness and Other Communication Disorders}.
\newblock Quick statistics about voice, speech, language, 2025.
\newblock Accessed: 2025-05-06.

\bibitem{wren2016prevalence}
Yvonne Wren, Laura~L. Miller, Tim~J. Peters, Alan Emond, and Sue Roulstone.
\newblock Prevalence and predictors of persistent speech sound disorder at eight years old: Findings from a population cohort study.
\newblock {\em Journal of Speech, Language, and Hearing Research}, 59(4):647--664, 2016.

\bibitem{cambridge_econ_impact}
James Law, Robert Rush, Ingrid Schoon, and Samantha Parsons.
\newblock The economic impact of low language ability in childhood.
\newblock {\em Language Development: Individual Differences in a Social Context}, pages 323--343, 2022.
\newblock doi:10.1017/9781108553567.018.

\bibitem{cock2009biopython}
Peter~JA Cock, Tiago Antao, Jeffrey~T Chang, Brad~A Chapman, Cymon~J Cox, Andrew Dalke, Iddo Friedberg, Thomas Hamelryck, Bartek Kauff, Bartosz Wilczynski, et~al.
\newblock Biopython: freely available python tools for computational molecular biology and bioinformatics.
\newblock {\em Bioinformatics}, 25(11):1422--1423, 2009.

\bibitem{blei2003latent}
David~M Blei, Andrew~Y Ng, and Michael~I Jordan.
\newblock Latent dirichlet allocation.
\newblock {\em Journal of Machine Learning Research}, 3:993--1022, 2003.

\bibitem{grootendorst2022bertopic}
Maarten Grootendorst.
\newblock Bertopic: Neural topic modeling with class-based tf-idf.
\newblock {\em arXiv preprint arXiv:2203.05794}, 2022.

\bibitem{zenodo_record}
Ziang Wang and Amir Aryani.
\newblock Topic modeling for childhood speech disorder literature (lda and bertopic), 2025.
\newblock Zenodo. doi:10.5281/zenodo.15429840.

\end{thebibliography}

\end{document}